# DREAMSAT-2.0: TOWARDS A GENERAL SINGLE-VIEW ASTEROID 3D RECONSTRUCTION

Santiago Diaz *, Xinghui Hu †, Josiane Uwumukiza ‡, Giovanni Lavezzi §, Victor Rodriguez-Fernandez ¶ and Richard Linares ‖

To enhance asteroid exploration and autonomous spacecraft navigation, we introduce DreamSat-2.0, a pipeline that benchmarks three state-of-the-art 3D reconstruction models—Hunyuan-3D, Trellis-3D, and Ouroboros-3D—on custom spacecraft and asteroid datasets. Our systematic analysis, using 2D perceptual (image quality) and 3D geometric (shape accuracy) metrics, reveals that model performance is domain-dependent. While models produce higher-quality images of complex spacecraft, they achieve better geometric reconstructions for the simpler forms of asteroids. New benchmarks are established, with Hunyuan-3D achieving top perceptual scores on spacecraft but its best geometric accuracy on asteroids, marking a significant advance over our prior work.

## INTRODUCTION

Three-dimensional (3D) reconstruction allows the generation of new viewpoints from one or more views of a scene, that is, to generate two-dimensional (2D) images from different angles. Within the space domain, this functionality can be very useful, particularly for generating 3D models and new perspectives of space objects such as spacecraft and asteroids. 3D reconstruction offers significant potential for enhancing asteroid exploration, mapping, and resource assessment missions. With 3D reconstruction, it is possible to accurately map both the general shape of an object and its important details, in order to later generate a complete 3D model. Whether for observational purposes or for sampling or approach operations, knowing the shape, size, orientation, and texture of these objects provides a better overall understanding in any mission targeting such a body. By generating new perspectives from limited observational data, 3D reconstruction can facilitate autonomous spacecraft navigation, enabling more precise asteroid approaches and improving hazard detection. Furthermore, 3D reconstruction can be used to create global texture maps, providing seamless and

The authors S. Diaz, X. Hu, J. Uwumukiza, and G. Lavezzi have equally contributed to this work.

*Undergraduate Student, School of Computer Systems Engineering, Universidad Politécnica de Madrid, Madrid, Spain, *sa.diaz@alumnos.upm.es*.

†Undergraduate Student, Department of Mathematics and Computer Science, Massachusetts Institute of Technology, Cambridge, MA 02139, USA, *xinghui@mit.edu*.

‡Undergraduate Student, Department of Computer Science, Wellesley College, Wellesley, MA 02481, USA, *ju100@wellesley.edu*.

§Research Scientist, Department of Aeronautics and Astronautics, Massachusetts Institute of Technology, Cambridge, MA 02139, USA, *glavezzi@mit.edu*.

¶Associate Professor, School of Computer Systems Engineering, Universidad Politécnica de Madrid, Madrid, Spain, *victor.rfernandez@upm.es*.

‖Associate Professor, Department of Aeronautics and Astronautics, Massachusetts Institute of Technology, Cambridge, MA 02139, USA, *linaresr@mit.edu*

complete visual representations of asteroid surfaces that aids to create Artificial Intelligence (AI) models used to detect and track the bodies. The potential for detailed shape and texture analysis can also inform resource assessment by estimating the volume and surface composition of asteroids from novel observations, and enable the selection of optimal sites for in-situ resource extraction or sampling.

Prior work on spacecraft-specific 3D reconstruction has largely relied on primitive-based, single-view pipelines. Early approaches used Convolutional Neural Networks (CNNs) to generate super quadric primitives but struggled with limited generalization and shape-pose ambiguity, even when enhanced with specialized loss functions [1, 2]. More recently, hybrid methods combine primitive-based CNNs with 3D Gaussian Splatting (3DGS) for faster, photorealistic results, yet they remain constrained by the fidelity of the initial primitive approximation and CNN-induced pose inaccuracies [3]. The broader field of 3D reconstruction is currently dominated by two key techniques: Neural Radiance Fields (NeRF) [4] and 3D Gaussian Splatting (3DGS) [5]. While numerous NeRF variants have improved performance in areas like rendering speed and quality [6, 7], both NeRF and the faster 3DGS paradigm fundamentally rely on multiple input views to achieve high-fidelity results. Key limitations in current 3D reconstruction techniques are the poor generalization across scenes of techniques such as NeRF or 3DGS, and the requirement for multiple views of an object or scene to accurately recreate it in 3D. However, thanks to the rise of large-scale datasets containing 3D models, such as Objaverse [8], significant progress has been made in single-view 3D reconstruction, enabling the development of novel approaches, such as Zero123 [9]. By leveraging diffusion-based generation, it is capable of synthesizing multiple views from a single input image. The Zero123-XL model, an extension of Zero123, has been trained on the ObjaverseXL [10] dataset using fine-tuning techniques. This approach improves its performance across a wider range of scenes. Building on these advances in single-view 3D reconstruction, recent methods have introduced diffusion-based processes to improve reconstruction fidelity. DreamGaussian [11], for example, adapts 3DGS to generative tasks. Despite these advances, which have clearly improved 3D reconstruction across various domains, there remains significant room for improvement when applying these techniques in the context of space. An example of this is DreamSat [12], which fine-tunes the Zero123-XL model on a custom spacecraft dataset. When integrated into the DreamGaussian framework, this approach yields efficient, high-fidelity reconstructions with improved performance compared to the baseline. Moreover, when it comes to generating 3D images from single-view asteroid imagery, existing approaches fall short of addressing the core challenge we aim to solve. Specifically, works such as those by Chen et al. [13], Givens [14], and Asteroid-NeRF [15] rely on multiple views of the same object for reconstruction, running counter to the single-view paradigm inherent in 3D reconstruction.

In this work, we introduce DreamSat-2.0, an enhanced pipeline for single-view 3D reconstruction. We upgrade the original DreamSat framework by integrating and fine-tuning three state-of-the-art 3D reconstruction large-scale models on two custom datasets of spacecraft and asteroids. First, we test DreamSat-2.0 on the original spacecraft dataset [12] to evaluate the improvements through a set of 2D perceptual and 3D geometric metrics. Then, DreamSat-2.0's capabilities are evaluated on a curated dataset of 3D asteroid models. Furthermore, we conduct a comparative analysis to evaluate the performance of the three reconstruction models, establishing new benchmarks for 3D generation in the space domain. Specifically, the models that have drawn our attention include Tencent's Hunyuan-3D-2.0 [16], Trellis-3D [17], and Ouroboros-3D [18]. Hunyuan-3D-2.0 combines diffusion-based geometry and texture generation to produce high-quality, detailed meshes from im-ages. Trellis-3D leverages a structured latent representation to accurately preserve both geometry and appearance across various formats, including Gaussians and meshes. Ouroboros-3D jointly trains view synthesis and reconstruction in a recursive diffusion loop, improving geometric consistency from limited input. These models are particularly promising for our task, as they are designed to generalize well from minimal views, aligning with the single-image constraint presented in this work. Our approach aims to adapt the capabilities of 3D reconstruction models and to leverage their generalization power to reconstruct a wide variety of asteroids. This would eliminate the need for case-specific training or individual close-range observations, while maintaining a high level of accuracy in capturing both global geometry and fine surface details.

The paper is organized as follows. First, we introduce the state-of-the-art 3D reconstruction models that are integrated into our DreamSat-2.0 pipeline. Subsequently, we describe the evaluation metrics, custom datasets used for our comparative analysis, and the DreamSat-2.0 architecture. Then, we discuss the experimental results for both spacecraft and asteroid reconstruction. Lastly, we provide final remarks to conclude the paper.

## STATE-OF-THE-ART 3D RECONSTRUCTION MODELS

In this section, we describe the three state-of-the-art models for single-view 3D generation that are included in DreamSat-2.0 and used in the comparison: Tencent's Hunyuan-3D-2.0 [16], Trellis-3D [17], and Ouroboros-3D [18].

### Hunyuan-3D-2.0

Hunyuan-3D 2.0 [16] adopts a two-stage generation pipeline, offering greater flexibility and control by separating the generation of 3D geometry (without texture) from the application of high-resolution texture to the generated or preexisting mesh. In this work, we focus on the first part of the pipeline: the generation of 3D geometry. This model is inspired by latent diffusion architectures and is designed to produce detailed 3D shapes from a single input image. The Hunyuan-3D architecture comprises the following:

- Hunyuan-3D-ShapeVAE (Shape Autoencoder): Its purpose is to compress the shape of a polygonal mesh into a sequence of tokens in a latent space. It uses "vector sets" as a compact neural representation for 3D shapes. The encoder employs an attention-based architecture, encoding point clouds sampled from the 3D surface. A key innovation is importance sampling, which concentrates points in high-frequency regions (such as edges and corners) to capture fine details and avoid artifacts. The encoder input includes 3D coordinates and normal vectors. The decoder reconstructs a 3D neural field from the latent input, predicting a Signed Distance Function (SDF), which can be converted into a triangle mesh using the Marching Cubes algorithm.
- Hunyuan-3D-DiT (Shape Diffusion Model): This is the core component responsible for shape generation. It adopts a flow-based diffusion network architecture with both dual-stream and single-stream blocks, enhancing the interaction between shape and image modalities. For conditioning, it uses a large pre-trained image encoder and pre-processes the input image to optimize effective resolution and reduce the negative influence of background elements. It is trained using a flow matching objective and employs an Euler ordinary differential equations solver during inference.

**Ouroboros-3D**

The Ouroboros-3D framework [18] operates through iterative cycles, where each cycle consists of a denoising step and a reconstruction step. This recursive nature allows the model to progressively refine the accuracy of the 3D reconstruction. The Ouroboros-3D architecture comprises the following:

- Multi-view Diffusion Generator: This component is responsible for generating a set of images from different viewpoints based on a single input image. It uses a video diffusion model; specifically, Stable Video Diffusion [19], adapted for this task.
- Feed-Forward Reconstruction Model: This module generates a 3D representation from the multiple views produced by the previous component. Ouroboros-3D uses the Large Multi-View Gaussian Model (LGM) [20] for 3D reconstruction, leveraging its efficiency in real-time rendering.
- 3D-Aware Feedback Mechanism: This is the core innovation of Ouroboros-3D. It allows the multi-view generative model to receive 3D geometric feedback from the previous reconstruction step to guide the next round of denoising. In practice, the reconstructed 3D model produces rendered color images and Canonical Coordinate Maps (CCMs). These are used as conditions to influence the denoising process of the multi-view generation. CCMs are preferred over other maps (such as depth or normals) because they encode normalized global vertex coordinates, offering a more robust view of the geometric relationships between views.
- Recursive Diffusion Process and Joint Training: Unlike two-stage methods that treat denoising independently, Ouroboros-3D generates an image prediction at every denoising step, which is then used for the next 3D reconstruction. The rendered maps from the resulting 3D model are fed back as conditions for the following denoising step. To bridge the gap between diffusion-generated images and training data, Ouroboros-3D implements joint training of both the multi-view diffusion model and the reconstruction model. During training, the reconstruction model processes images restored by the diffusion process instead of original images. This joint training not only improves reconstruction accuracy but also supervises the multi-view diffusion model, guiding it to generate more suitable images for few-shot reconstruction.

**Trellis-3D**

Trellis-3D [21] introduces a novel approach based on a Structured Latent Representation (SLAT), designed to efficiently capture both the geometry and appearance of 3D objects while allowing decoding into multiple 3D formats. A two-stage pipeline is used to generate the structured representations from a single image. The Trellis-3D architecture comprises the following:

- SLAT Encoding (Structured Latents Encoding): To convert a 3D model into the SLAT representation, images are first rendered from randomly sampled camera views around the object. These images are processed by a pre-trained DINOv2 [22] encoder, a powerful foundational vision model, to extract detailed feature maps. Finally, a Variational Autoencoder (VAE) encodes these features into structured latent spaces. This VAE handles voxel sparsity effectively by incorporating positional encodings to enhance local information interactions.
- SLAT Decoding: Each latent space representation is decoded into 32 Gaussian primitives with associated properties such as position, color, scale, opacity, and rotation, making them well-suited for efficient rendering.

- Sparse Structure Generation: The voxel layout is first generated using a VAE with 3D convolutional blocks. A transformer module (called GS) produces this layout, incorporating text or image conditioning through cross-attention mechanisms.
- Local Latent Generation: Given the active voxels, the associated latent representations are generated. Another transformer (GL), designed for sparse structures, performs this step using convolutional layers.

**EVALUATION METRICS**

Evaluating the quality of a 3D reconstruction is a multidimensional task that requires metrics that capture both visual fidelity and geometric accuracy. For this reason, we distinguish between metrics based on 2D renderings and those based on direct 3D comparisons. Table 1 reports a summary of the evaluation metrics.

2D perceptual metrics assess the visual accuracy of reconstructions by comparing rendered views of the reconstructed 3D model with reference images of the original model. They are useful for quantifying whether the model is visually consistent from various viewpoints. PSNR (Peak Signal-to-Noise Ratio) measures the ratio between the maximum possible power of a signal and the power of corrupting noise that affects its fidelity. It is a simple pixel-wise metric where higher values indicate lower pixel differences and thus better visual fidelity. Useful for evaluating distortion or noise introduced during reconstruction. Unlike PSNR, SSIM (Structural Similarity Index) [23] is designed to model human perception of image quality. It evaluates similarity between two images based on luminance, contrast, and structure. SSIM values closer to 1 indicate greater structural similarity and better perceived visual quality. LPIPS (Learned Perceptual Image Patch Similarity) uses a pre-trained neural network to extract features from images and compute the distance between them [24]. LPIPS has been shown to correlate much better with human judgments of perceptual similarity than PSNR or SSIM. Lower LPIPS values indicate higher perceptual similarity. It is especially useful for capturing subtle differences not evident to low-level metrics, and is crucial for realism evaluation.

3D geometric metrics directly compare the geometry of the reconstructed 3D model with that of the ground-truth model. They are essential for assessing structural and shape accuracy. 3D Volumetric IoU (Intersection over Union) measures the overlap between the volume of the reconstructed model and that of the ground-truth model. It is calculated as the ratio between the volume of their intersection and the volume of their union. To do this, both models are rescaled to the same size, aligned, and overlaid. A large number of random points is then sampled, and the IoU is computed as the ratio between the number of points inside both models (intersection) and the number of points inside at least one of them (union). This robust metric quantifies how well the reconstructed model captures the global shape and spatial extent of the real object. An IoU closer to 1 indicates a better volumetric match. Chamfer Distance (CD) measures the average distance from each point in one point cloud to its nearest neighbor in the other, and vice versa—each point cloud corresponding to the mesh of one of the 3D models. Specifically, it computes the mean squared distance from each point in one set to its nearest neighbor in the other, and sums it with the same operation in the opposite direction. A lower CD indicates that, on average, the points in one model are very close to those in the other, providing a good indication of global shape similarity. Hausdorff Distance (HD) measures the maximum distance from any point in one set to its closest point in the other, considering the maximum of these distances in both directions. Unlike CD, HD captures the "worst-case" deviation: it identifies the greatest distance from any point in one model to the other. A low value indicates that all points are near each other, ensuring that no large local discrepancies exist. It is useful for quantifying extreme cases of geometric dissimilarity.

Table 1: Summary of 3D Reconstruction Evaluation Metrics

| Metric | Description | Interpretation | Primary Use Case |
| --- | --- | --- | --- |
| *2D Perceptual Metrics (Render-based)* | | | |
| PSNR | Measures pixel-wise error between the rendered and reference images. | Higher is better | Quantifying low-level noise, distortion, and simple image fidelity. |
| SSIM | Evaluates similarity based on luminance, contrast, and structural information. | Closer to 1 is better | Measuring perceived visual quality that aligns better with human vision than PSNR. |
| LPIPS | Computes distance between deep features from a pre-trained neural network. | Lower is better | Capturing complex, perceptual similarity and evaluating overall realism. |
| *3D Geometric Metrics (Mesh-based)* | | | |
| IoU | Measures the overlap between the volumes of the reconstructed and ground-truth models. | Closer to 1 is better | Assessing global shape accuracy and spatial extent. |
| CD | Average nearest-neighbor distance between the point clouds of the two models. | Lower is better | Evaluating overall geometric similarity and global alignment. |
| HD | The maximum "worst-case" distance from any point in one model to the other. | Lower is better | Identifying the largest local deviations and quantifying extreme dissimilarities. |

## DATASETS PREPARATION

Two datasets are considered, consisting of collections of 3D models of spacecraft and asteroids. The original spacecraft dataset from [12] is adopted, characterized by 210 high-quality spacecraft models from National Aeronautics and Space Administration (NASA) [25], European Space Agency (ESA) [26], and Satellite Pose Estimation and 3D Reconstruction (SPE3R) [27, 28], which comprises 64 unique spacecraft 3D models, allowing for generalization across common spacecraft designs. This diverse set of models encompasses a wide range of spacecraft types, including satellites, space probes, and orbital stations. To prepare the dataset for both inference and fine-tuning of the reconstruction models, we use the tools from our custom framework to generate consistent views for each 3D model. These tools ensure that all views share the same format, distance, background, and other parameters, regardless of variations in the original models' size, position, or file format. The new dataset of 3D models of asteroids is obtained by combining two online databases, each containing a collection of 3D asteroid models. On the NASA Planetary Data System [29] a curated selection of 3D models of celestial bodies can be found, most of which are asteroids and comets. Although the number of models is not very large, 45 models in total, all of them are of high quality. The 3D Asteroid Catalogue [30] contains 1660 models, which can be categorized into three main types: lightcurve-based models, which are the most abundant but also the lowest in quality; radar-based models, which offer good quality, comparable to those from the previously mentioned website; and imagery-based models, which are the highest quality, as they are generated from direct images of the celestial bodies. Despite the predominance of lightcurve-based models, the dataset also includes a considerable number of radar-based and spacecraft imagery-based models. Figure 1 shows a comparison between the quality of lightcurve-, radar-, and imagery-based models.

To ensure a fair comparison with DreamSat-1.0 [12], we adopt its data splits. We partition the spacecraft dataset into a 190-model fine-tuning set and a 20-model test set, and the asteroid dataset into 95 models for fine-tuning and 20 for testing. The fine-tuning sets are used exclusively for adaptable models (e.g., Trellis-3D), while all models are benchmarked on the held-out test sets.

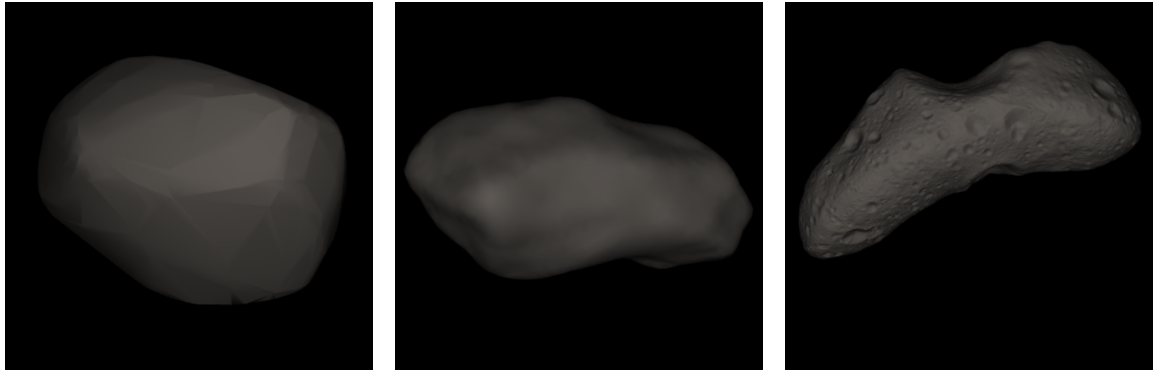

(a) Lightcurve-based model of asteroid (608) Adolfine.

(b) Radar-based model of asteroid (4769) Castalia.

(c) Imagery-based model of asteroid (433) Eros.

Figure 1: Comparison of lightcurve, radar and imagery-based asteroid models.

## DREAMSAT-2.0

In this section, we present the DreamSat-2.0 pipeline with model-agnostic evaluation framework.

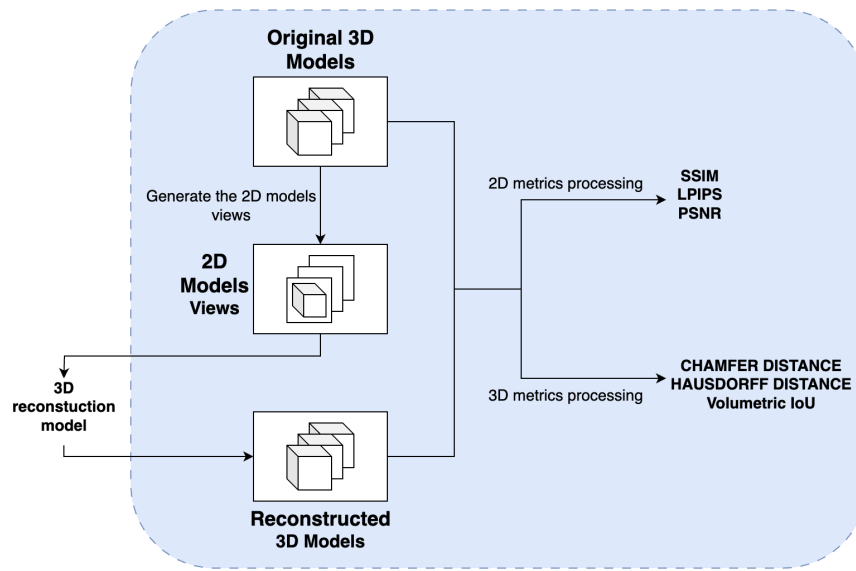

Figure 2: DreamSat-2.0 pipeline with evaluation framework.

### Architecture/Pipeline Design

The overall workflow is designed to be model-agnostic, modular, and fully automated, ensuring that all models, despite their architectural differences, are evaluated under the same standardized conditions. This flexibility is critical, as modern 3D reconstruction methods generate a wide range of outputs, from point clouds to polygonal meshes, textured or otherwise. As shown in Fig. 2, the pipeline supports various forms of output while maintaining consistent evaluation criteria. To begin using the system, only a set of original 3D models is required. Optionally, 2D views of each model can be provided. If not available, the framework can generate them automatically. These views are generated by creating a virtual camera to take a picture of the object in a controlled scene, using Trimesh and Pyrender to manage mesh loading, transformation, and rendering, entirely within a headless environment. This eliminates any reliance on third-party applications or Graphical User Interfaces (GUIs), such as Blender, allowing full automation in remote computing environments. Once the 2D images are available, they are fed into the designated 3D reconstruction model, which outputs a reconstructed 3D representation. The pipeline is independent from the model; the evaluation system only requires the output path to the reconstructed meshes. All subsequent stages (geometry alignment, view rendering, metric computation) are handled by the framework itself.

**Integration and Usage of Selected 3D Reconstruction Models**

For each model, its inference pipeline is executed separately for every input image across the evaluation datasets, which included 3D models of spacecraft and asteroids. These input images are synthetically generated under controlled conditions using the internal framework tools described previously, guaranteeing consistency in lighting, angle, and resolution. For all the models, we run the complete pipeline with texture generation disabled, resulting in clean polygonal meshes, independent of the format in which they are in. The three models selected for this study, Hunyuan-3D-2.0, Ouroboros-3D, and Trellis-3D, are evaluated under the same inference conditions using our model-agnostic evaluation framework.

For Hunyuan-3D-2.0, we employ its official open-source implementation, using the Hunyuan-3D-DiT-v2-0 checkpoint. This is the most powerful version of the model available for 3D shape reconstruction (excluding texture generation). For each input image in the evaluation datasets, the complete pipeline is executed with the texture generation stage disabled. This produced polygonal meshes in .glb format (i.e., GL Transmission Format Binary, a standardized file format used for 3D data), directly compatible with the evaluation and metric pipeline. The process is entirely automated within the evaluation framework. No fine-tuning is performed, as no training code or procedures are publicly available at the time of execution. A future extended study could explore re-training possibilities with the recently announced updates to the repository.

For Ouroboros-3D, the version used corresponds to the publicly released implementation of Ouroboros-3D, employing the original pre-trained weights that include both the multi-view diffusion generator and the reconstruction module (LGM). The full inference pipeline is run for each 2D input image. Ouroboros-3D produces 3DGS representations as output. These are post-processed and included in the evaluation suite using our unified pipeline, enabling the measurement of both geometric and perceptual performance. No domain-specific adaptation is applied to this model. The evaluation is conducted using its general version to assess default performance.

For Trellis-3D, we use its official open-source implementation, together with pre-trained weights supporting SLAT encoding and decoding. Both mesh decoders (for mesh generation) and 3DGS decoders are made available and used during evaluation where relevant. For each input image, Trellis-3D is executed to first generate its SLAT representation, followed by decoding using the DM mesh decoder, producing polygonal meshes for the 3D metric evaluation. Unlike the other models, Trellis-3D is fine-tuned specifically for this work. We finetune the sparse structure VAE separately for the spacecraft and asteroid datasets, allowing the model to specialize in the structural and visual characteristics unique to each domain. This adaptation aimed to assess the model's capacity for domain-specific improvements.

**Importance of a Model-Agnostic Evaluation Framework**

Modern 3D reconstruction models differ not only in architecture but also in output format and representation. A model-agnostic evaluation framework is fundamental to ensuring fair and unbiased comparison across methods. Without it, discrepancies in evaluation protocols (e.g., inconsistent camera angle or render resolution) can lead to misleading results. By using a centralized rendering and metric computation system, we ensure that all reconstructions—regardless of their generative pipeline—are evaluated with identical scenes, lighting, camera settings, and sampling strategies. This approach also supports large-scale and scalable benchmarking, facilitating the incorporation of future models into the study with minimal additional effort.

**RESULTS**

In this section, we report the 3D reconstruction results related to the spacecraft and asteroid dataset. Both training and inference of the models were done with NVIDIA RTX 4090 GPUs with the Python programming language. In addition, The Trellis-3D sparse structure VAE was finetuned with 3 GPUs, and ran at around 10,000 steps per hour with a batch size of 4 per GPU. We used the checkpoint saved at steps 300,000 and 600,000 to compute metrics for the finetuned Trellis-3D model.

**Spacecraft 3D Reconstruction**

A direct comparison of the 2D metrics (i.e., image quality) between Hunyuan-3D and Trellis-3D reveals several key points. For PSNR, Hunyuan-3D exhibits a higher mean value (20.93) compared to Trellis-3D (18.48). Its performance range is wider, reaching a significantly higher maximum (33.15 vs. 25.05), which suggests its ability to generate higher-fidelity reconstructions in certain cases. In terms of SSIM, both models achieve very high and similar mean values (0.92 for Hunyuan-3D and 0.91 for Trellis-3D), indicating high structural similarity in the generated images. Regarding LPIPS, Hunyuan-3D obtains a lower mean score (0.11) than Trellis-3D (0.15). Since a lower LPIPS value signifies greater perceptual similarity, Hunyuan-3D appears to generate images that are, on average, more aligned with human perception. The Ouroboros-3D model scores lower across these 2D metrics, with a mean PSNR of 18.98, SSIM of 0.91, and LPIPS of 0.14.

Regarding the 3D metrics (i.e., geometric accuracy), Hunyuan-3D performs significantly better on the IoU metric, with a mean value of 0.20 compared to a mean value of 0.12 for Trellis-3D and 0.12 for Ouroboros-3D. However, the significant variability is noteworthy, with a range spanning from 0.0001 (failure in overlap) to 0.80 (very high overlap). This indicates that the model's geometric accuracy is inconsistent on this dataset. For CD, Hunyuan-3D shows a lower mean distance (0.19) than Trellis-3D (0.66) and Ouroboros-3D (0.68), suggesting better overall alignment between the generated and ground-truth point clouds. On HD as well, Hunyuan-3D outperforms Trellis-3D by a large margin, with a mean value of 0.37 compared to a mean value of 0.51, which could suggest better handling of outliers. In this category, Ouroboros-3D also demonstrates lower geometric accuracy, with a mean IoU of 0.12 and higher (worse) mean CD and HD scores. This performance can be influenced by its output format; Ouroboros-3D generates a point cloud, which requires an intermediate conversion step (not included in its original framework) to create a mesh for rendering and evaluation. Further investigation is needed to understand whether that is affecting its performance.

Table 2: Results for spacecraft 3D reconstruction based on 2D and 3D metrics.

| Model | Metric | Mean | Max | Min |
|---|---|---|---|---|
| Hunyuan-3D | PSNR↑ | **20.93** | 33.15 | 14.87 |
| | SSIM↑ | **0.92** | 0.98 | 0.82 |
| | LPIPS↓ | **0.11** | 0.26 | 0.007 |
| Trellis-3D | PSNR↑ | 18.48 | 25.05 | 13.07 |
| | SSIM↑ | 0.91 | 0.96 | 0.85 |
| | LPIPS↓ | 0.15 | 0.27 | 0.05 |
| Ouroboros-3D | PSNR↑ | 18.98 | 24.80 | 14.93 |
| | SSIM↑ | 0.91 | 0.96 | 0.85 |
| | LPIPS↓ | 0.14 | 0.22 | 0.06 |
| Hunyuan-3D | IoU↑ | **0.20** | 0.80 | 0.0001 |
| | CD↓ | **0.19** | 18.31 | 0.0001 |
| | HD↓ | **0.37** | 3.44 | 0.04 |
| Trellis-3D | IoU↑ | 0.12 | 0.56 | 0.001 |
| | CD↓ | 0.66 | 11.53 | 0.0001 |
| | HD↓ | 0.51 | 2.85 | 0.03 |
| Ouroboros-3D | IoU↑ | 0.12 | 0.34 | 0.0005 |
| | CD↓ | 0.68 | 11.93 | 0.0007 |
| | HD↓ | 0.49 | 2.74 | 0.073 |

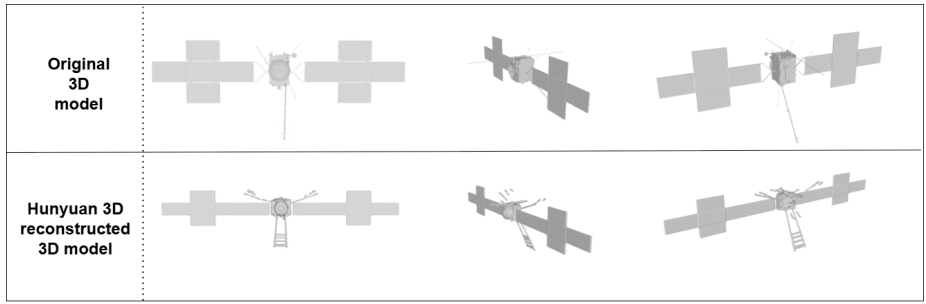

(a) Successful reconstruction showing consistent and symmetric geometry.

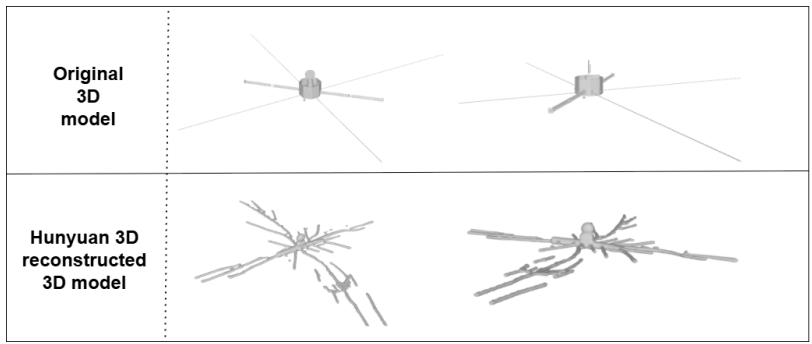

(b) Unsuccessful reconstruction showing structural inaccuracies.

Figure 3: Examples of successful (a) and unsuccessful (b) spacecraft reconstructions generated by Hunyuan-3D.

An example of 3D reconstructions generated by the Hunyuan-3D model is shown in Fig. 3a. Most reconstructions follow this general style: while they may lack very fine-grained details, the overall structure, components, and recognizable parts of the object are typically well preserved. In particular, many of the spacecraft reconstructions exhibit a clear representation of main modules, solar panels, and general symmetry. One noticeable trait of the reconstructions is their tendency toward symmetry, especially in cases where the input image does not reveal strong asymmetries. This may be due to the model's learned priors or inherent architectural biases favoring symmetric completions in ambiguous regions. However, not all outputs conform to this satisfactory reconstruction. In a small number of cases, the model fails to produce accurate or plausible shapes, generating outputs that are not representative of spacecraft structures. These failure cases are relatively rare, but when they do occur, they result in reconstructions with unclear geometry or unrealistic proportions, as shown in Fig. 3b.

**Asteroid 3D Reconstruction**

Hunyuan-3D shows a different behavior compared to its performance on the spacecraft dataset. In the 2D metrics, the mean PSNR is lower (18.75 vs. 20.93), the SSIM is slightly lower (0.91 vs. 0.92), and the LPIPS is higher (0.13 vs. 0.11), indicating worse perceptual quality. This suggests that reconstructing asteroid images poses a greater challenge for the model in terms of image fidelity. The performance range is also notably narrower, pointing towards more consistent, albeit lower-quality, results. When compared to the other models, Hunyuan-3D outperforms Trellis-3D across all 2D metrics. Ouroboros-3D presents a more complex case, achieving the highest mean PSNR score of 19.94, yet falling slightly behind Hunyuan-3D on SSIM and LPIPS.

In contrast, Hunyuan-3D's 3D metrics show a marked improvement. The IoU is considerably higher and more stable, with a mean of 0.56 and a minimum value of 0.28, surpassing Ouroboros-3D's 0.27 and Trellis-3D's 0.35. This indicates a much more reliable and consistent geometric accuracy. Similarly, both the CD (mean 0.016) and HD (mean 0.20) are significantly lower (better) than on the spacecraft dataset. This reinforces the observation that the model achieves a more precise and robust geometric reconstruction in the asteroid domain. However, for surface-based metrics, Ouroboros-3D achieves the best CD (0.011), indicating a highly accurate point cloud representation. Hunyuan-3D leads on HD, though all models remain competitive, suggesting that the point cloud to mesh conversion step in Ouroboros-3D may be impacting its volumetric IoU while its raw point cloud output excels in geometric precision.

A distinct pattern can be observed in the 3D reconstructions of asteroids. Since most asteroids resemble roughly a spherical shape, their 3D geometry tends to be easier to generalize. This simplicity in structure allows reconstruction models like Hunyuan-3D to produce accurate outputs even with limited visual information, often resulting in significantly higher scores in geometric metrics such as IoU. In general, the reconstructions do not rely on fine-grained structural cues as in the case of spacecraft, which makes the reconstruction task inherently more forgiving. As a result, the outputs are more uniform and geometrically plausible, even when the model receives only a single view as input. An example of a high-quality reconstruction given only a single view of an asteroid can be seen in Fig. 4a. Despite this tendency toward high-quality outputs, there are still occasional reconstructions in which Hunyuan-3D underperforms. Nevertheless, even these less accurate results are not overly poor, especially considering the challenging nature of reconstructing 3D geometry from a single input view. An example of one of the worst reconstructions in the asteroid dataset, in terms of geometric fidelity and similarity to the original shape, is shown in Fig. 4b.

Table 3: Results for asteroid 3D reconstruction based on 2D and 3D metrics.

| Model | Metric | Mean | Max | Min |
|---|---|---|---|---|
| Hunyuan-3D | PSNR↑ | 18.75 | 22.42 | 16.35 |
|  | SSIM↑ | **0.91** | 0.94 | 0.85 |
|  | LPIPS↓ | **0.13** | 0.17 | 0.07 |
| Trellis-3D | PSNR↑ | 16.09 | 20.98 | 14.51 |
|  | SSIM↑ | 0.90 | 0.94 | 0.88 |
|  | LPIPS↓ | 0.19 | 0.23 | 0.08 |
| Ouroboros-3D | PSNR↑ | **19.94** | 27.77 | 17.70 |
|  | SSIM↑ | 0.90 | 0.93 | 0.88 |
|  | LPIPS↓ | 0.13 | 0.17 | 0.08 |
| Hunyuan-3D | IoU↑ | **0.56** | 0.87 | 0.28 |
|  | CD↓ | 0.016 | 0.05 | 0.001 |
|  | HD↓ | **0.20** | 0.34 | 0.05 |
| Trellis-3D | IoU↑ | 0.35 | 0.75 | 0.23 |
|  | CD↓ | 0.02 | 0.067 | 0.001 |
|  | HD↓ | 0.24 | 0.41 | 0.06 |
| Ouroboros-3D | IoU↑ | 0.27 | 0.39 | 0.17 |
|  | CD↓ | **0.011** | 0.02 | 0.001 |
|  | HD↓ | 0.21 | 0.29 | 0.17 |

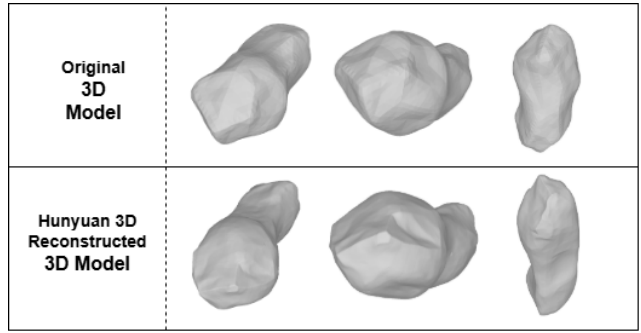

(a) A good reconstruction that geometrically approximates the original object.

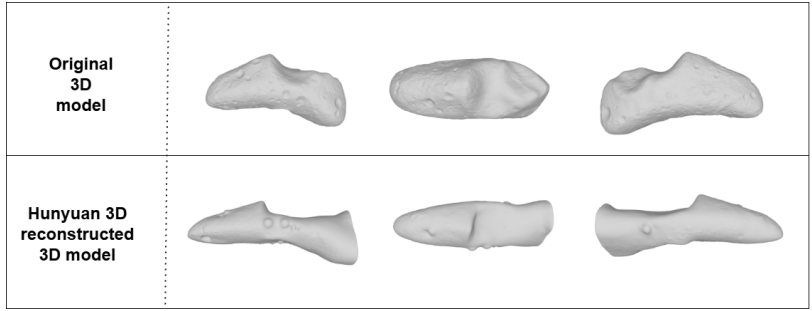

(b) A poor reconstruction that deviates from the original shape.

Figure 4: Examples of high-quality (a) and lower-quality (b) asteroid reconstructions from Hunyuan-3D.

**Analysis of Reconstruction Errors**

Elaborating on the previously discussed reconstruction failures, we performed a manual inspection of Hunyuan-3D outputs to pinpoint characteristic errors. While the model can produce high-fidelity results, it struggles with specific geometric features and component relationships. Figure 5 showcases a collection of outputs that illustrate this spectrum. For instance, Figures 5b and 5d are high-quality reconstructions that exhibit none of the aforementioned issues, demonstrating the model's capability under ideal conditions. However, subtle errors can appear even in otherwise accurate models, e.g. in Fig. 5f which, despite its high quality, fails to correctly connect the solar panels to the satellite's main body. Finally, Figure 5h exemplifies a more severe failure related to the misinterpretation of long, thin structures. In this instance, the antennas are excessively elongated, a flaw that significantly distorts the main body's geometry and compromises the overall structural integrity, resulting in a poor reconstruction. This limitation shows that errors in fine-grained details can propagate and lead to a major distortion of the object. Interestingly, the results in Table 4 highlight a discrepancy between the perceptual (2D) and geometric (3D) metrics, particularly for spacecraft with complex, sparse components like antennas (e.g., Objects 4). Although object 4's reconstruction is visually less accurate and scores lower on 2D metrics, it achieves strong scores on 3D metrics. This occurs because the model tends to replace fine-grained, difficult-to-reconstruct features with simpler geometric shapes that happen to occupy the same general volume. Moreover, 3D metrics are highly sensitive to the pre-processing steps of the models (e.g., alignment, scaling, and positioning).

| Metric | Spacecraft | | | | Asteroid | | | |
| --- | --- | --- | --- | --- | --- | --- | --- | --- |
| | Obj. 1 | Obj. 2 | Obj. 3 | Obj. 4 | Obj. 1 | Obj. 2 | Obj. 3 | Obj. 4 |
| PSNR↑ | **21.80** | 19.50 | 20.10 | 17.90 | **20.06** | 19.18 | 19.43 | 18.00 |
| SSIM↑ | 0.91 | **0.92** | 0.90 | 0.89 | **0.929** | 0.927 | 0.86 | 0.89 |
| LPIPS↓ | **0.11** | 0.12 | 0.12 | 0.15 | 0.12 | **0.11** | 0.12 | 0.15 |
| IoU↑ | 0.02 | 0.07 | **0.18** | 0.17 | **0.77** | 0.55 | 0.49 | 0.39 |
| CD↓ | 0.001 | 0.02 | 0.04 | **0.0008** | 0.01 | 0.03 | **0.004** | 0.02 |
| HD↓ | 0.13 | 0.33 | 0.40 | **0.09** | 0.18 | 0.31 | **0.18** | 0.26 |

Table 4: Metrics for each reconstructed spacecraft and asteroid object.

Similarly, Figure 6 presents a series of asteroid reconstructions that highlight different levels of success and failure. Analyzing these models reveals common issues, such as the inability to capture complete geometry from a single viewpoint and a tendency to distort the object's overall shape. The reconstruction in Fig. 6b serves as a high-quality benchmark, representing a successfully reconstructed object without significant errors. In contrast, Figures 6d and 6f illustrate a more subtle limitation; although these are generally good reconstructions, they demonstrate that relying on a single view prevents the model from accurately capturing hidden geometric details. Despite this incompleteness, they remain effective reconstructions given the input constraints. A more severe failure is evident in Fig. 6h, where the model excessively elongates or "stretches" the asteroid, significantly altering its fundamental geometry and resulting in a poor reconstruction that fails to preserve the object's true shape. This suggests that the model may misinterpret perspective, leading to major structural distortions. Table 4 lists the performance metrics for each reconstructed object shown in Fig. 6.

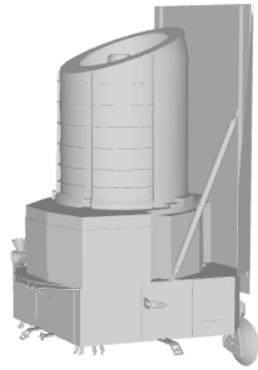

(a) Ground truth (Object 1)

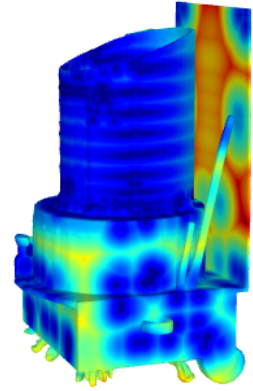

(b) Reconstruction (Object 1)

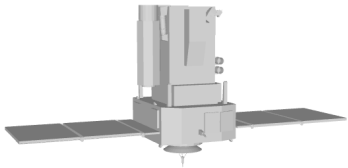

(c) Ground truth (Object 2)

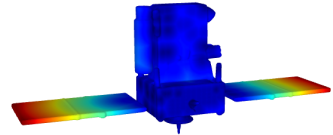

(d) Reconstruction (Object 2)

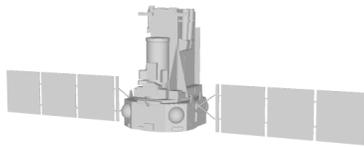

(e) Ground truth (Object 3)

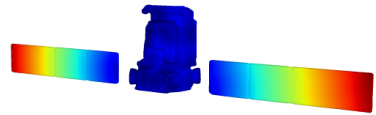

(f) Reconstruction (Object 3)

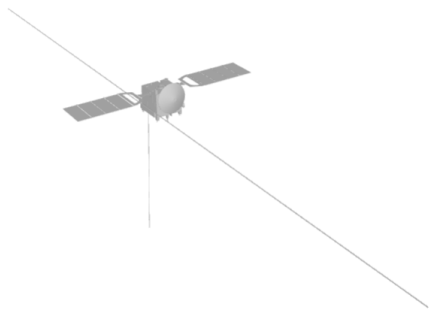

(g) Ground truth (Object 4)

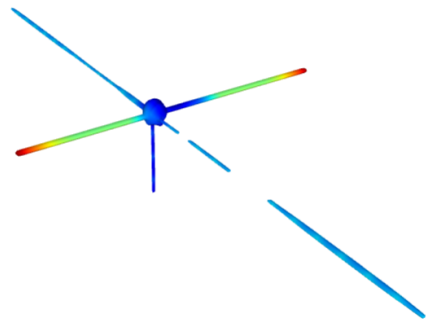

(h) Reconstruction (Object 4)

**Figure 5**: Examples of spacecraft reconstruction outputs. Each row shows a ground-truth model (left) and its corresponding reconstruction (right), where the normalized color scale, which is independent for each object, represents an error gradient from blue (lower error/distance) to red (higher error/distance), based on the IoU metric.

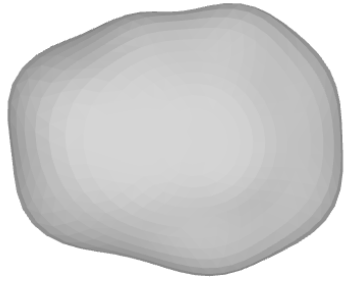
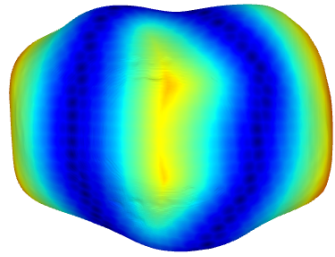

(a) Ground truth (Object 1)　　　　　　　　(b) Reconstruction (Object 1)

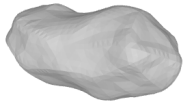
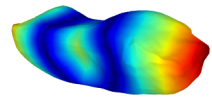

(c) Ground truth (Object 2)　　　　　　　　(d) Reconstruction (Object 2)

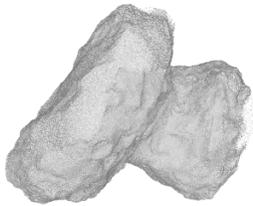
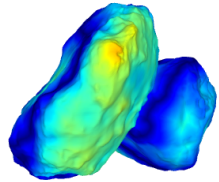

(e) Ground truth (Object 3)　　　　　　　　(f) Reconstruction (Object 3)

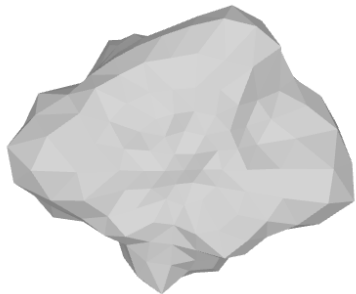
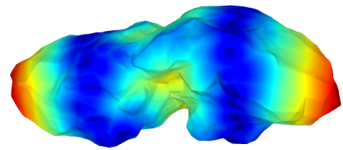

(g) Ground truth (Object 4)　　　　　　　　(h) Reconstruction (Object 4)

**Figure 6**: Examples of asteroid reconstruction outputs. Each row shows a ground-truth model (left) and its corresponding reconstruction (right), where the normalized color scale, which is independent for each object, represents an error gradient from blue (lower error/distance) to red (higher error/distance), based on the IoU metric.

**Trellis-3D fine-tuned**

On the spacecraft dataset, Table 5 reports a slight improvement to all metrics after 300,000 steps of finetuning. For the 2D metrics, the mean PSNR is slightly higher (18.51 vs 18.48), the SSIM is slightly higher (0.919 vs. 0.918), and the LPIPS is lower (0.1511 vs 0.1513), indicating better perceptual quality. There is also an increase in mean IoU (0.1214 vs 0.1213) and a decrease in mean CD (0.6611 vs 0.6612) and mean HD (0.5119 vs 0.5129). After 600,000 steps of finetuning, the PSNR (mean 17.68), SSIM (mean 0.9161) and LPIPS (mean 0.1618) metrics imply worse performance in visual quality, even compared to the vanilla Trellis-3D model with no fine-tuning. This could be a sign of overfitting in the training of the model, which is likely to happen when training is done with a small set of data. Training other parts of the model such as the SLaT encoder and decoder could help to mitigate the problem by allowing the entire model to adapt to the finetuning dataset and align the changes across the different parts. On the 3D metrics, the mean IoU is lower (0.09 vs 0.1213) and the mean CD is higher (0.667 vs 0.6611) when compared to the vanilla Trellis-3D model, suggesting generally worse geometric accuracy of the generated pointcloud of a the finetuned model compared to the ground truth. However, the mean HD (0.50) is lower compared to both the vanilla and the 300,000-step finetuned models, suggesting a better handling of outliers with a longer period of finetuning.

Table 5: Results for spacecraft 3D reconstruction on different stages of the finetuned Trellis-3D model based on 2D and 3D metrics.

| Finetuning Steps | Metric | Mean | Max | Min |
| --- | --- | --- | --- | --- |
| 0 | PSNR↑ | 18.48 | 25.05 | 13.07 |
|   | SSIM↑ | 0.918 | 0.96 | 0.85 |
|   | LPIPS↓ | 0.1513 | 0.27 | 0.05 |
| 300,000 | PSNR↑ | **18.51** | 25.24 | 13.07 |
|   | SSIM↑ | **0.919** | 0.96 | 0.85 |
|   | LPIPS↓ | **0.151** | 0.27 | 0.05 |
| 600,000 | PSNR↑ | 17.68 | 24.64 | 13.25 |
|   | SSIM↑ | 0.91 | 0.96 | 0.85 |
|   | LPIPS↓ | 0.16 | 0.27 | 0.05 |
| 0 | IoU↑ | 0.1213 | 0.56 | 0.001 |
|   | CD↓ | 0.6612 | 11.53 | 0.0001 |
|   | HD↓ | 0.5129 | 2.85 | 0.03 |
| 300,000 | IoU↑ | **0.1214** | 0.55 | 0.001 |
|   | CD↓ | **0.6611** | 11.53 | 0.0001 |
|   | HD↓ | 0.5119 | 2.85 | 0.03 |
| 600,000 | IoU↑ | 0.09 | 0.39 | 0.001 |
|   | CD↓ | 0.667 | 11.62 | 0.0001 |
|   | HD↓ | **0.50** | 2.80 | 0.03 |

On the asteroid dataset, Table 6 reports a clear improvement in the perceptual quality of the model, with an increase in PSNR (17.07 vs 16.09) and SSIM (0.91 vs 0.90) and a decrease in LPIPS (0.175 vs 0.19) at 600,000 steps of finetuning compared to the vanilla Trellis-3D model. This suggests that finetuning allows Trellis-3D to reconstruct images with better image fidelity and are more aligned to human visual perception. For the 3D metrics, there is an increase in IoU (0.40 vs 0.35) as well, implying a larger overlap in the volume of the generated asteroid model compared to the ground truth. However, there is an increase in both CD (mean 0.03 vs 0.02) and HD (mean 0.27 vs 0.24), implying that the point clouds of the generated and original models differ more after finetuning.

Table 6: Results for asteroid 3D reconstruction on different stages of the finetuned Trellis-3D model based on 2D and 3D metrics.

| Finetuning Steps | Metric | Mean | Max | Min |
|---|---|---|---|---|
| 0 | PSNR↑ | 16.09 | 20.98 | 14.51 |
|   | SSIM↑ | 0.901 | 0.94 | 0.88 |
|   | LPIPS↓ | 0.19 | 0.23 | 0.08 |
| 300,000 | PSNR↑ | 16.73 | 21.73 | 15.15 |
|   | SSIM↑ | 0.908 | 0.93 | 0.88 |
|   | LPIPS↓ | 0.179 | 0.22 | 0.08 |
| 600,000 | PSNR↑ | **17.07** | 23.83 | 15.20 |
|   | SSIM↑ | **0.91** | 0.93 | 0.89 |
|   | LPIPS↓ | **0.175** | 0.22 | 0.07 |
| 0 | IoU↑ | 0.35 | 0.75 | 0.23 |
|   | CD↓ | **0.02** | 0.06 | 0.001 |
|   | HD↓ | **0.24** | 0.41 | 0.06 |
| 300,000 | IoU↑ | 0.35 | 0.76 | 0.02 |
|   | CD↓ | 0.03 | 0.08 | 0.001 |
|   | HD↓ | 0.26 | 0.44 | 0.06 |
| 600,000 | IoU↑ | **0.40** | 0.80 | 0.04 |
|   | CD↓ | 0.03 | 0.10 | 0.002 |
|   | HD↓ | 0.27 | 0.44 | 0.06 |

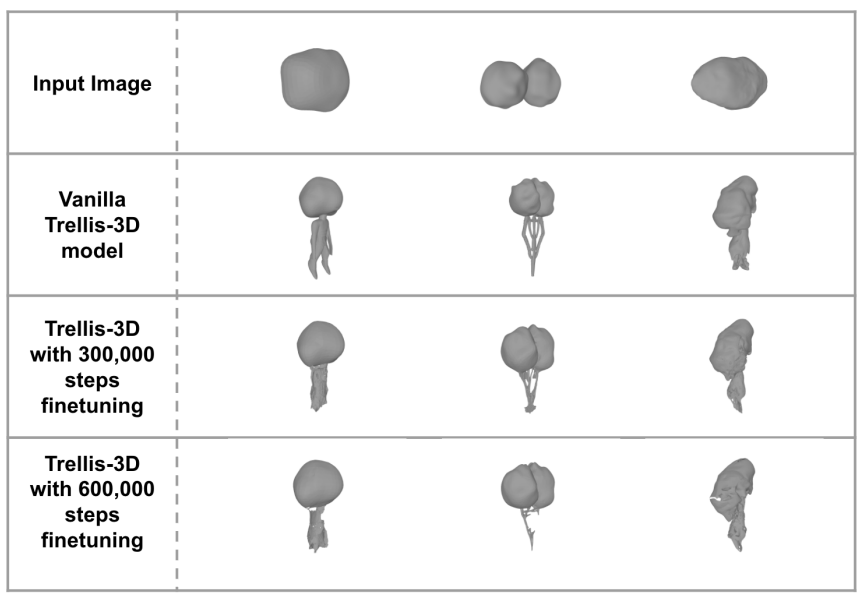

Figure 7: Example of asteroid reconstruction from Trellis-3D.

During 3D reconstruction for asteroid models, Trellis-3D tends to imbue its generated models with qualities of humanoids or more common objects, as can be seen in Fig. 7. This is likely an artifact of the original training and evaluation dataset of Trellis-3D, which contains models from ObjaverseXL and Toys4K. This could explain the difference in performance of the finetuned Trellis-3D on 2D and 3D metrics: by finetuning from this baseline, Trellis-3D may be able to optimize for clearer images from fixed viewpoints by distorting parts of the baseline 3D model, at the cost of compromising the geometric accuracy. By comparing the reconstructions in Fig. 7, the generated 3D model seems to be converging towards the geometric shape of the ground truth with increased number of steps for finetuning. It remains to see if an improvement in 3D metrics could eventually be obtained with sufficient data and over a longer period of training.

## CONCLUSION

In this study, we introduced DreamSat-2.0, an enhanced pipeline to rigorously benchmark the performance of three state-of-the-art models—Hunyuan-3D, Trellis-3D, and Ouroboros-3D—on custom datasets of spacecraft and asteroids. Our analysis reveals a significant domain-dependency in model performance. For instance, Hunyuan-3D excelled in perceptual quality on complex spacecraft, establishing a new and robust benchmark with a mean PSNR of 20.93 that marks a substantial leap from the relative gains shown in our prior work [12]. Conversely, on the geometrically simpler asteroid dataset, Hunyuan-3D delivers superior and more stable geometric accuracy with a high mean IoU of 0.5619. This discrepancy highlights that while intricate details challenge geometric consistency, the simpler forms of asteroids allow for more stable volumetric reconstruction. For future work, a valuable direction is to systematically investigate this relationship between object complexity and model performance to better understand the generalization capabilities of these architectures. Furthermore, due to computational constraints, we were only able to partially fine-tune the Trellis-3D model; fully fine-tuning all models on these domain-specific datasets remains a promising avenue to potentially unlock even greater performance. In conclusion, this work establishes a comprehensive performance baseline for large-scale 3D reconstruction models in the space domain. It underscores the necessity of domain-specific evaluation to select the appropriate tools for a given task. By providing a clear, quantitative framework, the DreamSat-2.0 pipeline advances the tools available for critical applications like Space Domain Awareness, rendezvous and proximity operations, and asteroid exploration, contributing to the space domain.

## CODE AVAILABILITY

The code for this work can be accessed on GitHub (https://github.com/ARCLab-MIT/space-nvs).

## ACKNOWLEDGMENTS

Research was partially supported by the Madrid Government (Comunidad de Madrid-Spain) under the Multiannual Agreement 2023-2026 with Universidad Politecnica de Madrid in the Line A, Emerging PhD researchers, and by the Department of the Air Force Artificial Intelligence Accelerator and was accomplished under Cooperative Agreement Number FA8750-19-2-1000. The views and conclusions contained in this document are those of the authors and should not be interpreted as representing the official policies, either expressed or implied, of the Department of the Air Force or the U.S. Government. The U.S. Government is authorized to reproduce and distribute reprints for Government purposes notwithstanding any copyright notation herein.

# REFERENCES

[1] T. H. Park, E. Bates, and S. D'Amico, "Improving Zero-Shot Abstraction of Unknown Spacecraft 3D Shape as Primitive Assembly," *International Workshop on Satellite Constellations and Formation Flying*, 2024.

[2] E. Bates and S. D'Amico, "Removing ambiguities in concurrent monocular single-shot spacecraft shape and pose estimation using a deep neural network," *2025 AAS GNC conference (Feb 2025)*, 2025.

[3] P. F. Huc, E. Bates, and S. D'Amico, "Fast Learning of Non-Cooperative SpaceCraft 3D Models Through Primitive Initialization," *AAS/AIAA Astrodynamics Specialist Conference, Boston, MA*, 2025.

[4] B. Mildenhall, P. P. Srinivasan, M. Tancik, J. T. Barron, R. Ramamoorthi, and R. Ng, "NeRF: Representing Scenes as Neural Radiance Fields for View Synthesis," *European conference on computer vision*, Springer, 2020, pp. 405–421.

[5] B. Kerbl, G. Kopanas, T. Leimkühler, and G. Drettakis, "3D Gaussian Splatting for Real-Time Radiance Field Rendering," *ACM Transactions on Graphics (TOG)*, Vol. 42, No. 4, 2023.

[6] Y. Zhao, S. Wu, J. Zhang, S. Li, C. Li, and Y. Lin, "Instant-NeRF: Instant On-Device Neural Radiance Field Training via Algorithm-Accelerator Co-Designed Near Memory Processing," *2023 60th ACM/IEEE Design Automation Conference (DAC)*, 2023, pp. 1–6.

[7] J. T. Barron, B. Mildenhall, D. Verbin, P. P. Srinivasan, and P. Hedman, "Mip-NeRF 360: Unbounded Anti-Aliased Neural Radiance Fields," *2022 IEEE/CVF Conference on Computer Vision and Pattern Recognition (CVPR)*, 2022, pp. 5460–5469, 10.1109/CVPR52688.2022.00542.

[8] M. Deitke, D. Schwenk, J. Salvador, L. Weihs, O. Michel, E. VanderBilt, L. Schmidt, K. Ehsani, A. Kembhavi, and A. Farhadi, "Objaverse: A Universe of Annotated 3D Objects," *2023 IEEE/CVF Conference on Computer Vision and Pattern Recognition (CVPR)*, 2023, pp. 13142–13153.

[9] R. Liu, R. Wu, B. V. Hoorick, P. Tokmakov, S. Zakharov, and C. Vondrick, "Zero-1-to-3: Zero-Shot One Image to 3D Object," *2023 IEEE/CVF International Conference on Computer Vision (ICCV)*, 2023, pp. 9264–9275, 10.1109/ICCV51070.2023.10276580.

[10] M. Deitke, R. Liu, M. Wallingford, H. Ngo, O. Michel, A. Kusupati, A. Fan, C. Laforte, V. Voleti, S. Y. Gadre, E. VanderBilt, A. Kembhavi, C. Vondrick, G. Gkioxari, K. Ehsani, L. Schmidt, and A. Farhadi, "Objaverse-Xl: A Universe of 10M+ 3D Objects," *Proceedings of the 37th International Conference on Neural Information Processing Systems (NeurIPS)*, Red Hook, NY, USA, Curran Associates Inc., 2024.

[11] J. Tang, J. Ren, H. Zhou, Z. Liu, and G. Zeng, "DreamGaussian: Generative Gaussian Splatting for Efficient 3D Content Creation," *International Conference on Learning Representations (ICLR)*, Vol. abs/2309.16653, 2023. arXiv:2309.16653.

[12] N. Mathihalli, A. Wei, G. Lavezzi, P. Mun Siew, V. Rodriguez-Fernandez, H. Urrutxua, and R. Linares, "DreamSat: Towards a General 3D Model for Novel View Synthesis of Space Objects," *75th International Astronautical Congress 2024*, Milan, Italy, International Astronautical Federation, 10 2024.

[13] H. Chen, X. Hu, K. Willner, Z. Ye, F. Damme, P. Gläser, Y. Zheng, X. Tong, H. Hußmann, and J. Oberst, "Neural Implicit Shape Modeling for Small Planetary Bodies from Multi-View Images Using a Mask-Based Classification Sampling Strategy," *ISPRS Journal of Photogrammetry and Remote Sensing*, Vol. 212, 2024, pp. 122–145.

[14] M. W., "Appliying neural radiance fields to asteroid shape modeling," 2024.

[15] S. Chen, B. Wu, H. Li, Z. Li, and Y. Liu, "Asteroid-NeRF: A deep-learning method for 3D surface reconstruction of asteroids," , Vol. 687, July 2024, p. A278, 10.1051/0004-6361/202450053.

[16] T. H. Team, "Hunyuan3D 2.0: Scaling Diffusion Models for High Resolution Textured 3D Assets Generation," 2025.

[17] J. Xiang, Z. Lv, S. Xu, Y. Deng, R. Wang, B. Zhang, D. Chen, X. Tong, and J. Yang, "Structured 3D Latents for Scalable and Versatile 3D Generation," *arXiv preprint arXiv:2412.01506*, 2024.

[18] H. Wen, Z. Huang, Y. Wang, X. Chen, Y. Qiao, and L. Sheng, "Ouroboros3D: Image-to-3D Generation via 3D-aware Recursive Diffusion," *arXiv preprint arXiv:2406.03184*, 2024.

[19] A. Blattmann, T. Dockhorn, S. Kulal, D. Mendelevitch, M. Kilian, D. Lorenz, Y. Levi, Z. English, V. Voleti, A. Letts, V. Jampani, and R. Rombach, "Stable Video Diffusion: Scaling Latent Video Diffusion Models to Large Datasets," 2023.

[20] J. Tang, Z. Chen, X. Chen, T. Wang, G. Zeng, and Z. Liu, "LGM: Large Multi-View Gaussian Model for High-Resolution 3D Content Creation," 2024.

[21] J. Xiang, Z. Lv, S. Xu, Y. Deng, R. Wang, B. Zhang, D. Chen, X. Tong, and J. Yang, "Structured 3D Latents for Scalable and Versatile 3D Generation," *arXiv preprint arXiv:2412.01506*, 2024.

[22] X. Song, X. Xu, and P. Yan, "General Purpose Image Encoder DINOv2 for Medical Image Registration," 2024.

[23] Z. Wang, A. Bovik, H. Sheikh, and E. Simoncelli, "Image quality assessment: from error visibility to structural similarity," *IEEE Transactions on Image Processing*, Vol. 13, No. 4, 2004, pp. 600–612, 10.1109/TIP.2003.819861.

[24] R. Zhang, P. Isola, A. A. Efros, E. Shechtman, and O. Wang, "The Unreasonable Effectiveness of Deep Features as a Perceptual Metric," 2018.

[25] NASA 3D Resources. https://nasa3d.arc.nasa.gov/.

[26] ESA Science Satellite Fleet. https://scifleet.esa.int/.

[27] T. H. Park and S. D'Amico, "Rapid Abstraction of Spacecraft 3D Structure from Single 2D Image," *AIAA SCITECH 2024 Forum*, 2024, 10.2514/6.2024-2768.

[28] SPE3R: Synthetic Dataset for Satellite Pose Estimation and 3D Reconstruction. https://purl.stanford.edu/pk719hm4806.

[29] NASA Planetary Data System. https://sbn.psi.edu/pds/shape-models/.

[30] 3D Asteroid Catalogue. https://3d-asteroids.space/asteroids/.